\documentclass[letterpaper,10pt,conference]{ieeeconf}

\IEEEoverridecommandlockouts
\usepackage[utf8]{inputenc}
\usepackage[T1]{fontenc}
\usepackage{amsmath}
\usepackage{amssymb}
\usepackage{amsfonts}
\usepackage{graphicx}
\usepackage{booktabs}
\usepackage{float}
\usepackage{array}
\usepackage{cite}
\usepackage{url}
\usepackage[hidelinks,bookmarks=false]{hyperref}
\makeatletter
\newcommand{\paperlinksfont}{\normalfont\@IEEEabskeysecsize\bfseries}
\makeatother

\newcommand{\normdbltable}{%
  \footnotesize
  \setlength{\tabcolsep}{4pt}
  \renewcommand{\arraystretch}{1.08}
}

\title{\LARGE \bfseries MPCoT: Reward-Guided Multi-Path Latent Reasoning for Test-Time Scalable Vision-Language-Action}
\author{Boyang Zhang$^{1}$ and Lianlei Shan$^{2}$%
\thanks{\raggedright $^{1}$Department of Electrical and Computer Engineering, Boston University, Boston, MA 02215, USA.
\texttt{theostnc@bu.edu}}%
\thanks{\raggedright $^{2}$Department of Computer Science, Tsinghua University, Beijing 100084, China.
\texttt{shanlianlei18@mails.ucas.edu.cn}}%
}

\hypersetup{
pdftitle={MPCoT: Reward-Guided Multi-Path Latent Reasoning for Test-Time Scalable Vision-Language-Action},
pdfauthor={Boyang Zhang; Lianlei Shan},
pdfsubject={Vision-language-action and multi-path latent reasoning},
pdfkeywords={robot learning; vision-language-action; test-time scaling}
}

\begin{document}
\maketitle

\begin{abstract}
Vision-Language-Action (VLA) policies remain brittle in long-horizon control, where one-pass action decoding offers limited inference-time deliberation.
Explicit chain-of-thought adds reasoning depth but incurs token-generation latency and an indirect text-to-action interface.
We propose \emph{MPCoT}, a reward-guided multi-path latent reasoning framework with configurable depth $K$ and width $M$.
MPCoT initializes $M$ latent hypotheses, refines them for $K$ weight-tied steps, and softly aggregates them before action decoding.
A training-only path-preference objective combines expert-trajectory consistency, frozen Qwen3-VL progress scores, and endpoint-consistency feedback to supervise the path scorer.
Under matched protocols on LIBERO and CALVIN, MPCoT improves long-horizon performance; ablations support the contributions of depth, width, soft aggregation, and reward supervision.
On five real-world ALOHA Mini bimanual tasks, average success increases from 71.3\% to 82.0\% over the matched OpenVLA-OFT baseline.
These results support latent deliberation as a means of improving execution while preserving the action interface and generating no reasoning tokens.
\end{abstract}
\begingroup
\paperlinksfont
\noindent\hspace*{\parindent}Video: {\urlstyle{same}\url{https://youtu.be/24cqXXB5DIw}}\par
\noindent\hspace*{\parindent}Code: {\urlstyle{same}\url{https://github.com/EDGSCOUT/MPCoT}}\par
\endgroup
\smallskip

\section{INTRODUCTION}

Vision-Language-Action (VLA) policies unify perception, language grounding, and action generation for robot manipulation \cite{rt1,palme,rt2}.
Open frameworks such as OpenVLA and OpenVLA-OFT support reproducible evaluation on benchmarks including LIBERO \cite{openvla,openvla_oft,libero}.
Yet reliable long-horizon execution remains difficult, particularly with compositional instructions and uncertain observations.
A one-pass action decoder offers little opportunity to reconsider a control hypothesis, allowing early errors to compound across later action chunks.

This exposes a tension between reasoning depth and control efficiency.
Explicit chain-of-thought (CoT) can improve multi-step reasoning in language models and VLA control \cite{cotprompt,cotvla}, but intermediate traces add latency and complicate the interface to continuous actions.
Continuous CoT, latent compression, SoftCoT, and recurrent-depth VLA instead motivate deliberation without decoded intermediate text \cite{coconut,codi,softcot,rdvla}.
Here, ``latent reasoning'' denotes iterative refinement and aggregation of continuous control hypotheses, rather than symbolic reasoning or decoded rationales.
The question is how to allocate inference compute between refining one hypothesis and maintaining alternatives, while preserving the action interface.

Depth refines a hypothesis; width preserves alternatives. We evaluate their success gains against added policy-query latency.

\begin{figure*}[!t]
  \centering
  \includegraphics[width=0.93\textwidth]{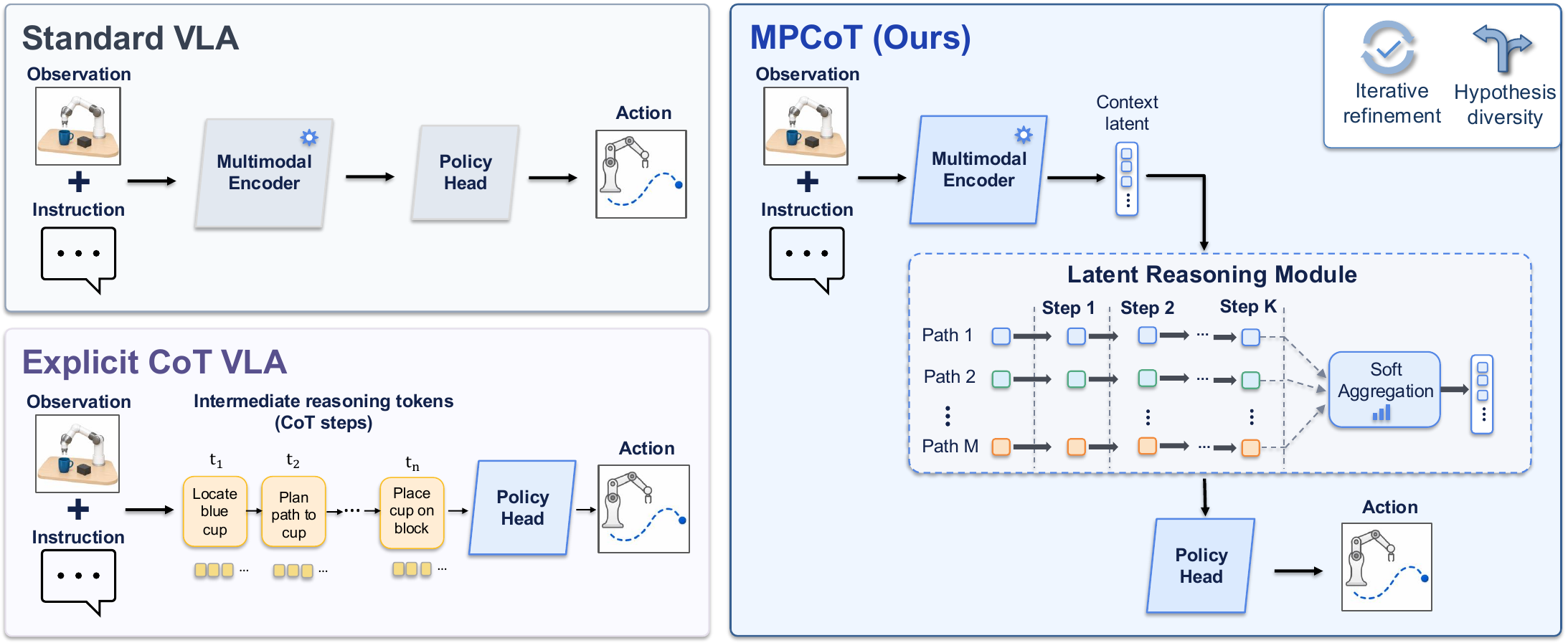}
  \caption{Conceptual comparison between standard VLA, explicit CoT VLA, and MPCoT. MPCoT keeps deliberation in continuous latent space, enabling multi-step/multi-path reasoning without changing the action interface or generating reasoning tokens.}
  \label{fig:method_overview}
\end{figure*}

We propose \emph{MPCoT}, a multi-path latent reasoning module for VLA decision making.
Given an observation and instruction, MPCoT refines $M$ hypotheses from the same grounded context for $K$ weight-tied steps.
A learned scorer softly aggregates these hypotheses before the unchanged action head.
Unlike single-path refinement, MPCoT retains alternatives and learns execution-aware preferences through training-only reward supervision.
At deployment, $(K,M)$ controls refinement depth and active hypothesis count, without reward evaluation or reasoning-token generation.
Fig.~\ref{fig:method_overview} contrasts standard VLA, explicit CoT VLA, and MPCoT.

\noindent\textbf{Contributions.}
The main contributions are fourfold:
\begingroup
\setlength{\itemsep}{0.15em}
\begin{itemize}
  \item We frame VLA reasoning as an embodied test-time compute allocation problem between high-cost explicit reasoning and shallow one-pass control.
  \item We introduce MPCoT, a recurrent multi-path latent reasoning module with explicit depth/width controls $(K,M)$, zero reasoning tokens, and an unchanged action interface.
  \item We introduce reward-guided path preference learning to align latent branch scores with downstream execution quality using trajectory consistency, frozen Qwen3-VL progress, and endpoint-consistency feedback.
  \item We validate MPCoT on LIBERO, CALVIN, and five real-world ALOHA Mini tasks, demonstrating improved long-horizon and real-world sequential execution.
\end{itemize}
\endgroup

\section{RELATED WORK}

\paragraph{Open VLA policies and benchmarks.}
Language-grounded robot policies have evolved toward open VLA backbones and efficient fine-tuning \cite{saycan,rt1,palme,rt2,openvla,openvla_oft}.
Methods such as $\pi_0$, UnifiedVLA, FLOWER, VLA-Adapter, and AVA-VLA improve VLA policies through action modeling, architecture design, and adaptation \cite{pi0,unifiedvla,flower_vla,vla_adapter,ava_vla}.
MPCoT instead studies inference-compute allocation within OpenVLA-OFT, adding latent deliberation before the existing action head.

\paragraph{Reasoning-augmented VLA and latent test-time compute.}
CoT-VLA, TraceVLA, and WorldVLA expose intermediate deliberation through text or visual traces, at the cost of intermediate-generation overhead \cite{cotvla,tracevla,worldvla}.
Continuous and compressed reasoning avoids decoded intermediate text \cite{coconut,codi,ccot,heima,tokenassorted}, while SoftCoT projects assistant-generated soft thoughts into the target LLM \cite{softcot}.
Closest to MPCoT, recurrent-depth VLA studies repeated updates along a single latent trajectory \cite{rdvla}, while broader test-time-compute work examines accuracy--latency trade-offs \cite{thinkingspeed}.
MPCoT combines depth and width through reward-guided aggregation. Its hypotheses share the backbone, refiner, and action head, without inference-time rollout.


\section{METHOD}
\label{sec:method}

\begin{figure*}[!t]
  \centering
  \includegraphics[width=0.93\textwidth]{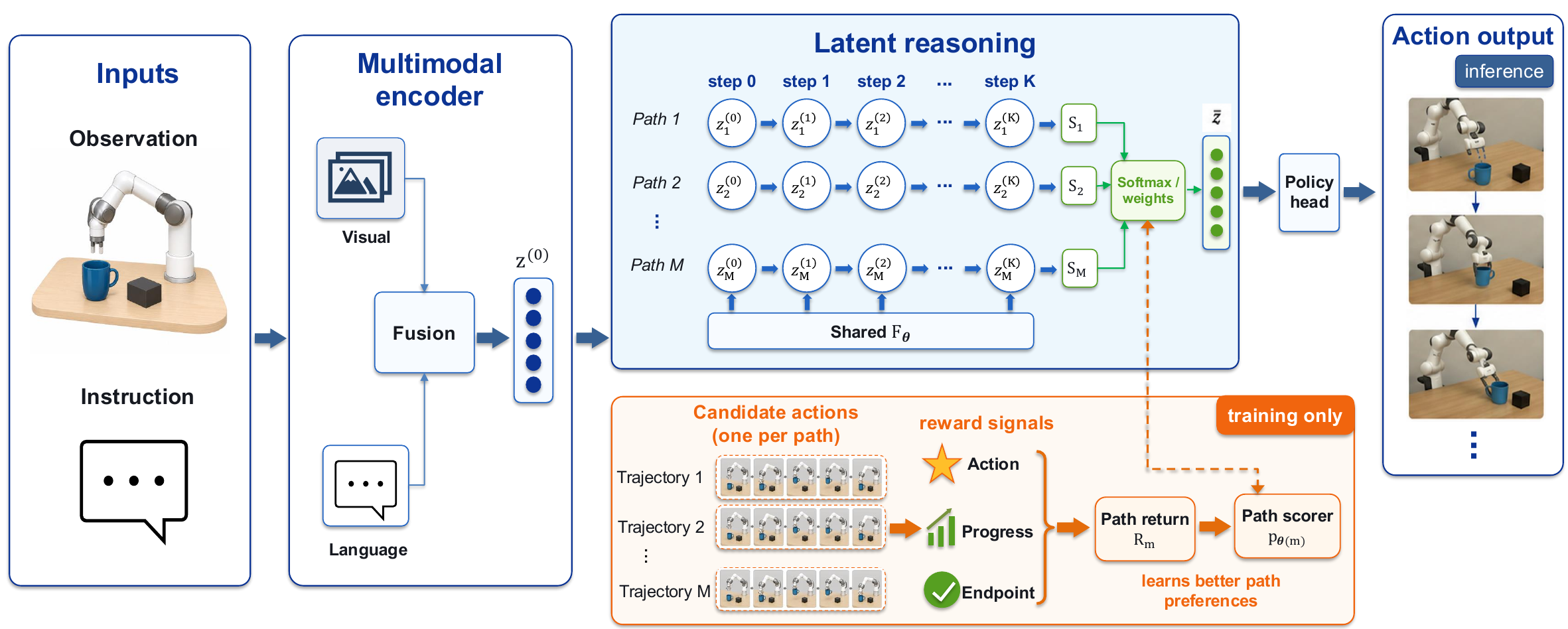}
  \caption{MPCoT architecture with training-time path supervision. Multiple latent branches are refined for $K$ shared steps, scored, and softly aggregated before action decoding. During training only, candidate action branches receive expert-trajectory action consistency, frozen Qwen3-VL progress scores, and endpoint-consistency feedback; this reward branch is removed at inference.}
  \label{fig:multipath_detail}
\end{figure*}

\noindent\textbf{Overview.}
Fig.~\ref{fig:multipath_detail} details the MPCoT architecture and its training-time path supervision.
At each control step, MPCoT initializes $M$ hypotheses, applies $K$ shared refinement steps, and aggregates the branches using learned soft weights.
Intermediate states remain continuous \cite{coconut,codi,ccot,heima,tokenassorted}; only the aggregated latent is decoded into an action chunk at inference.
During training, individual branches are also decoded to obtain reward labels for path-preference supervision.

\subsection{Problem Setup and Design Targets}
At each control step $t$, the policy receives observation $o_t$ and instruction $l$ and predicts action chunk $a_t$.
We apply a latent reasoning module $\mathcal{R}_{\theta}$ before the policy decoder $\pi_0$:
\begin{equation}
\bar{z}_t = \mathcal{R}_{\theta}(o_t,l;K,M),
\qquad
a_t \sim \pi_0(a\mid \bar{z}_t),
\label{eq:problem_setup}
\end{equation}
We instantiate $\pi_0$ with OpenVLA-OFT \cite{openvla_oft}, keeping its action interface fixed within each embodiment.
The module exposes depth and width controls while limiting parameter growth through weight tying.
Setting $(K,M)=(0,1)$ bypasses the reasoning module for direct decoding; $M=1$ with $K>0$ gives single-path latent refinement.
\subsection{Backbone Interface and Latent Hypothesis Initialization}
Let $f_v(\cdot)$, $f_l(\cdot)$, and $f_{\mathrm{mm}}(\cdot)$ denote the visual encoder, language encoder, and multimodal fusion block.
The fused latent context is
\begin{equation}
c_t = f_{\mathrm{mm}}\!\big(f_v(o_t),f_l(l)\big), \qquad c_t\in\mathbb{R}^d.
\label{eq:mm_context}
\end{equation}
Here $d$ follows the OpenVLA-OFT latent width.
All paths share the grounded context, so their differences arise after perception and language encoding.
To create $M$ parallel reasoning hypotheses, we initialize
\begin{equation}
z_{t,m}^{(0)} = c_t + e_m + \epsilon_{t,m}, \quad m=1,\dots,M,
\label{eq:path_init}
\end{equation}
where $e_m=\sigma_e u_m$, $u_m\stackrel{\mathrm{i.i.d.}}{\sim}\mathcal{N}(0,I_d)$, is a sampled code with fixed scale $\sigma_e=0.02$.
Codes are resampled per training sequence, held fixed across its control steps and refinement iterations, and never optimized.
At inference, a generator seeded with 7 supplies the first $M$ codes from the same distribution and scale, retaining earlier codes as $M$ increases; the training jitter $\epsilon_{t,m}$ is zero.
Increasing $M$ from two to four adds two initial states processed by the trained shared refiner and scorer, without new path-specific parameters.
\subsection{Weight-Tied Multi-Step Latent Refinement}
For each path, we unroll a shared refinement operator $F_{\theta}$ for $K$ steps:
\begin{equation}
r_{t,m}^{(k)} = F_{\theta}\!\left([\mathrm{LN}(z_{t,m}^{(k-1)});c_t]\right),
\label{eq:delta_update}
\end{equation}
\begin{equation}
z_{t,m}^{(k)} = z_{t,m}^{(k-1)} + r_{t,m}^{(k)}, \quad k=1,\dots,K.
\label{eq:path_update}
\end{equation}
Eqs.~\eqref{eq:delta_update}--\eqref{eq:path_update} implement residual updates with $F_{\theta}$ shared across steps and paths.
Paths run in parallel; $c_t$ retains the original perception-language context at every step.

\subsection{Reward-Guided Path Preference Learning}
\label{sec:path_preference}
After $K$ refinement steps, MPCoT obtains $M$ candidate latent branches $z_{t,m}^{(K)}$.
A lightweight scorer assigns path logits and soft selection probabilities:
\begin{equation}
s_{t,m}=h_{\omega}\!\left(z_{t,m}^{(K)}\right), \quad
p_{\omega}(m\mid o_t,l)=\frac{\exp(s_{t,m}/\tau)}{\sum_{j=1}^{M}\exp(s_{t,j}/\tau)} .
\label{eq:path_score_weight}
\end{equation}
The same probabilities serve as aggregation weights:
\begin{equation}
\bar{z}_{t}^{(K)}=\sum_{m=1}^{M}p_{\omega}(m\mid o_t,l)\,z_{t,m}^{(K)}.
\label{eq:latent_agg}
\end{equation}

For training-only path supervision, the unchanged policy head decodes each refined branch into a candidate action chunk:
\begin{equation}
a_{t,m}\sim \pi_0(a\mid z_{t,m}^{(K)}).
\label{eq:path_action_decode}
\end{equation}
Candidate branches are evaluated offline without simulator rollout. The first signal measures trajectory-level consistency with the expert chunk.
For the simulation interface, $H=8$ and $C_h(a)=\sum_{q=1}^{h}a_q$ for action step $h\in\{1,\ldots,H\}$; the absolute-joint counterpart is specified below. The action reward is
\begin{equation}
r_{t,m}^{\mathrm{act}}
=-\frac{1}{H}\sum_{h=1}^{H}
\left\|C_h(a_{t,m})-C_h(a_t^\ast)\right\|_2^2.
\label{eq:path_reward_act}
\end{equation}

\begin{table*}[t]
\centering
\normdbltable
\caption{OpenVLA-OFT and MPCoT implementation settings.}
\label{tab:exp_params}
\begin{tabular*}{\textwidth}{@{\extracolsep{\fill}}lp{0.31\textwidth}lp{0.31\textwidth}@{}}
\toprule
Item & Value & Item & Value \\
\midrule
Backbone & OpenVLA-OFT (7B) & Fine-tuning & LoRA r=32; parallel decoding; L1 continuous action \\
Sim. inputs & 2 views + language + proprioception, 224$\times$224 & Sim. inference & 8-step chunk (predict 8, execute 8 open-loop) \\
Optimizer & LR $5\!\times\!10^{-4}$; decay to $5\!\times\!10^{-5}$ at 100K; BS=64 & Train steps & 150K (S/O/L); 50K (Goal) \\
Training candidates & $K=3$, $M=2$ & Main inference & $K=5$, $M=4$ \\
Largest test budget & $K=7$, $M=6$ & $\tau$; $\lambda_{\mathrm{div}},\gamma_{\mathrm{div}}$ & 0.7; 0.05, $d/4$ \\
Progress evaluator & Qwen3-VL (frozen; training only) & Reward weights & $\beta_a=\beta_p=\beta_s=1$ \\
$\lambda_{\mathrm{RL}}$; baseline & 1.0; path average & $\beta_K,\beta_M$ & 4, 3 \\
\midrule
Real robot & ALOHA Mini (bimanual) & Real inputs & 3 RGB views + language + proprioception, 224$\times$224 \\
Real action & 14-D absolute joint action & Real control & 25 Hz; 25-step action chunk \\
Real tasks & 5 & Real evaluation & 30 trials/task/method \\
\bottomrule
\end{tabular*}
\end{table*}

Frozen Qwen3-VL independently predicts each candidate's progress from current images, instruction, task context, robot state, and candidate trajectory, without future rollout images or expert actions. Simulation uses front- and wrist-camera views.

For simulation progress evaluation, outputs are denormalized to 7-D relative end-effector actions in \texttt{robot\_base}: translation in meters, rotation increments (\texttt{droll}, \texttt{dpitch}, \texttt{dyaw}) in radians, and gripper commands ($-1$: close; $1$: open). The fixed prompt states these conventions. The eight-step input also includes \texttt{cumulative\_translation}, the final step's gripper command (\texttt{final\_gripper\_state}), and current end-effector position, orientation, and gripper state. Natural-language goal conditions derive from LIBERO BDDL predicates or CALVIN task IDs and oracle logic, not observed success; redundant goal fields are omitted.

For real-world training, forward kinematics converts each candidate's 25 native 14-D absolute joint actions into per-arm end-effector position and orientation sequences, retaining gripper commands separately. Qwen3-VL scores the full trajectory with current observations, robot state, and instruction. This training-only conversion preserves the native control interface and does not execute candidates.

For real-robot consistency, set $H=25$ and $C_h(a)=\mathcal{N}(a_h)$ in Eqs.~\eqref{eq:path_reward_act} and~\eqref{eq:path_reward_end}.
$\mathcal{N}$ applies the policy's fixed per-dimension action normalization and gripper encoding to candidates and aligned expert targets.
Action consistency compares all 25 targets; endpoint consistency compares the final target, without accumulating absolute joint angles.
The scale $0.05$ is dimensionless here: endpoint consistency measures target-configuration agreement, not Cartesian distance or an observed endpoint.

The prompt scores expected task completion, not motion magnitude, considering action order, plausible contact or grasp, and gripper state. Anchors are 0 (no useful progress or likely failure), 0.25 (small correct movement), 0.5 (meaningful partial completion), 0.75 (near completion), and 1 (expected success-condition satisfaction). The compact user prompt is:
\begin{quote}
\small
TASK INSTRUCTION: \textit{instruction}. SUCCESS CONDITION: \textit{condition}. CURRENT ROBOT STATE: \textit{state}. CANDIDATE TRAJECTORY: \textit{actions}. Estimate the expected task progress after executing the complete candidate trajectory from the attached current observation images. Return only: \texttt{\{"progress": <number between 0.0 and 1.0>\}}.
\end{quote}
Deterministic decoding uses no sampling, temperature 0, and at most 32 new tokens, returning only the JSON progress field. Scores are clipped to $[0,1]$; parsing failures are invalid. Valid scores $p_{t,m}$ are standardized within the same-state candidate set:
\begin{equation}
r_{t,m}^{\mathrm{prog}}
=\frac{p_{t,m}-\mu_t}{\sigma_t+\epsilon},\quad
\mu_t=\frac{1}{M}\sum_{j=1}^{M}p_{t,j},
\label{eq:path_reward_prog}
\end{equation}
where $\sigma_t$ is the population standard deviation and $\epsilon=10^{-6}$. Since candidates share the current state, no separate Qwen3-VL call scores it. Normalized rewards are detached. The third signal measures offline endpoint consistency in the corresponding action representation:
\begin{equation}
\begin{aligned}
e_{t,m}&=\left\|C_H(a_{t,m})-C_H(a_t^\ast)\right\|_2,\\
r_{t,m}^{\mathrm{end}}&=\exp\!\left(-e_{t,m}/0.05\right).
\end{aligned}
\label{eq:path_reward_end}
\end{equation}
Expert actions supervise consistency, not Qwen3-VL; endpoint agreement is not observed task success. The evaluator is fixed across reward ablations. Its scalar interface permits replacement without architecture or inference changes; performance across evaluators remains untested. Reward-label computation is detached.
Together these signals yield the composite branch reward:
\begin{equation}
r_{t,m}
=
\beta_a r_{t,m}^{\mathrm{act}}
+
\beta_p r_{t,m}^{\mathrm{prog}}
+
\beta_s r_{t,m}^{\mathrm{end}} .
\label{eq:path_reward}
\end{equation}
We convert rewards into a discounted return, path-average baseline, and advantage:
\begin{equation}
\begin{aligned}
R_{t,m}&=\sum_{\ell=t}^{T}\gamma^{\ell-t}r_{\ell,m},\\
b_t&=\frac{1}{M}\sum_{j=1}^{M}R_{t,j},\qquad
A_{t,m}=R_{t,m}-b_t.
\end{aligned}
\label{eq:path_return_advantage}
\end{equation}
The scorer is trained with the advantage-weighted path-selection objective
\begin{equation}
\mathcal{L}_{\mathrm{RL}}
=-\mathbb{E}_{m\sim p_{\omega}}
\left[\operatorname{sg}(A_{t,m})\log p_{\omega}(m\mid o_t,l)\right],
\label{eq:loss_rl}
\end{equation}
where $\operatorname{sg}(\cdot)$ denotes stop-gradient.
This objective trains path preferences alongside behavior cloning.
Rewards, returns, baselines, and advantages are detached, and the reward objective updates only the scorer.
For this loss, gradients stop at the scorer inputs; behavior cloning retains gradients through both aggregation weights and latent branches.
At inference, MPCoT uses the learned scorer and Eq.~\eqref{eq:latent_agg}, without candidate rollout, Qwen3-VL queries, or access to reward labels.

\subsection{Action Decoding and Training Objective}
The unchanged policy head decodes the aggregated latent into an action chunk, retaining the 8-step horizon in simulation:
\begin{equation}
a_t \sim \pi_0(a\mid \bar{z}_t^{(K)}).
\label{eq:action_decode}
\end{equation}
The behavior-cloning and diversity terms are
\begin{equation}
\mathcal{L}_{\mathrm{BC}}=\mathbb{E}_{(o_t,l,a_t^\ast)}
\!\left[\ell\!\left(\pi_0(\cdot\mid\bar{z}_t^{(K)}),a_t^\ast\right)\right],
\label{eq:loss_bc}
\end{equation}
\begin{equation}
\mathcal{L}_{\mathrm{div}}=
\frac{1}{M(M-1)}\sum_{m\neq j}
\exp\!\left(-\frac{\|z_{t,m}^{(K)}-z_{t,j}^{(K)}\|_2^2}{\gamma_{\mathrm{div}}}\right).
\label{eq:loss_div}
\end{equation}
Here $\gamma_{\mathrm{div}}$ is the diversity scale, distinct from the return discount $\gamma$; $\mathcal{L}_{\mathrm{div}}=0$ for $M=1$.
The complete end-to-end objective is
\begin{equation}
\mathcal{L}=\mathcal{L}_{\mathrm{BC}}+\lambda_{\mathrm{RL}}\mathcal{L}_{\mathrm{RL}}+\lambda_{\mathrm{div}}\mathcal{L}_{\mathrm{div}}.
\label{eq:loss_total}
\end{equation}
Training budgets govern candidate supervision and compute dropout ($\hat{K}\leq K$, $\hat{M}\leq M$).
Inference independently varies shared-refiner depth $K$ and hypothesis count $M$ using the same checkpoint; aggregation preserves the decoder's $d$-dimensional input.

\begin{table*}[t]
\centering
\normdbltable
\caption{Comparison on LIBERO. (a) One policy for all 4 suites. (b) One policy per suite.}
\label{tab:libero}
\begin{tabular*}{\textwidth}{@{\extracolsep{\fill}}lccccc@{}}
\toprule
\multicolumn{6}{l}{\textit{(a) One policy for all four suites.}} \\
\midrule
Method & Spatial SR (\%) & Object SR (\%) & Goal SR (\%) & Long SR (\%) & Average SR (\%) \\
\midrule
TraceVLA~\cite{tracevla} & 84.6 & 85.2 & 75.1 & 54.1 & 74.8 \\
WorldVLA~\cite{worldvla} & 87.6 & 96.2 & 83.4 & 60.0 & 81.8 \\
$\pi_0$~\cite{pi0} & 96.8 & 98.8 & 95.8 & 85.2 & 94.2 \\
$\pi_0$-FAST~\cite{pifast} & 96.4 & 96.8 & 88.6 & 60.2 & 85.5 \\
UnifiedVLA~\cite{unifiedvla} & 95.4 & 98.8 & 93.6 & 94.0 & 95.5 \\
OpenVLA-OFT~\cite{openvla_oft} & 97.7 & 98.0 & 96.1 & 95.3 & 96.8 \\
AVA-VLA~\cite{ava_vla} & 97.4 & 99.4 & 97.4 & 97.6 & 98.0 \\
\textbf{MPCoT (Ours)} & \textbf{98.2} & \textbf{99.7} & \textbf{98.6} & \textbf{98.9} & \textbf{98.9} \\
\bottomrule
\end{tabular*}

\vspace{0.6em}

\begin{tabular*}{\textwidth}{@{\extracolsep{\fill}}lccccc@{}}
\toprule
\multicolumn{6}{l}{\textit{(b) One policy per suite.}} \\
\midrule
Method & Spatial SR (\%) & Object SR (\%) & Goal SR (\%) & Long SR (\%) & Average SR (\%) \\
\midrule
OpenVLA~\cite{openvla} & 84.7 & 88.4 & 79.2 & 53.7 & 76.5 \\
SpatialVLA~\cite{spatialvla} & 88.2 & 89.9 & 78.6 & 55.5 & 78.1 \\
CoT-VLA~\cite{cotvla} & 87.5 & 91.6 & 87.6 & 69.0 & 83.9 \\
NORA~\cite{nora_vla} & 92.2 & 95.4 & 89.4 & 74.6 & 87.9 \\
PD-VLA~\cite{pdvla} & 95.5 & 96.7 & 94.9 & 91.7 & 94.7 \\
UniVLA~\cite{univla} & 96.5 & 96.8 & 95.6 & 92.0 & 95.2 \\
OpenVLA-OFT~\cite{openvla_oft} & 97.6 & 98.4 & 97.9 & 94.5 & 97.1 \\
FLOWER~\cite{flower_vla} & 97.5 & 99.1 & 96.1 & 94.9 & 96.9 \\
VLA-Adapter~\cite{vla_adapter} & 97.8 & 99.2 & 97.2 & 95.0 & 97.3 \\
RIPT-VLA~\cite{ript_vla} & 99.0 & 98.6 & 98.6 & 93.8 & 97.5 \\
AVA-VLA~\cite{ava_vla} & 99.2 & 99.6 & 98.2 & 96.2 & 98.3 \\
\textbf{MPCoT (Ours)} & \textbf{99.5} & \textbf{99.8} & \textbf{99.0} & \textbf{97.8} & \textbf{99.0} \\
\bottomrule
\end{tabular*}
\end{table*}

\subsection{Architecture and Complexity}
$F_{\theta}$ is a lightweight weight-tied residual MLP with LayerNorm, expansion ratio 4, GELU, and dropout 0.1. The path scorer $h_{\omega}$ is a two-layer MLP over final path states. Both modules are shared across paths.
Let $C_{\mathrm{base}}$ denote backbone cost, $C_{\mathrm{ref}}$ per-step refinement cost, and $C_{\mathrm{score}}$ per-path scoring cost. For fixed $(K,M)$,
\begin{equation}
C_{\mathrm{test}}\approx C_{\mathrm{base}}+KM\,C_{\mathrm{ref}}+M\,C_{\mathrm{score}}.
\label{eq:complexity}
\end{equation}
Refinement scales with $KM$ compute and $\mathcal{O}(Md)$ latent-state memory, with shared parameters.
MPCoT adds about 2.7\% parameters relative to the backbone; Table~\ref{tab:scaling_form} reports measured latency.
All quantitative results use fixed configurations. As an optional deployment extension, previous-step uncertainty $u_{t-1}$ can allocate compute through
\begin{equation}
\begin{aligned}
K_t&=\min\!\left(K_{\max},K_{\min}+\lfloor\beta_Ku_{t-1}\rfloor\right),\\
M_t&=\min\!\left(M_{\max},M_{\min}+\lfloor\beta_Mu_{t-1}\rfloor\right).
\end{aligned}
\label{eq:adaptive_budget}
\end{equation}
At $t=0$, the controller uses the default budget. This controller is not used in the reported results, and Qwen3-VL is never queried at inference.










\section{EXPERIMENTS}
\label{sec:experiments}

\subsection{Experimental Setup, Metrics, and Protocol}
\label{sec:exp_setup}
Simulation experiments use OpenVLA-OFT with official evaluation code and splits.
Unless noted, training and evaluation follow OpenVLA-OFT defaults~\cite{openvla_oft}, including the 8-step action interface.
Main results use inference $(K,M)=(5,4)$ unless specified; Table~\ref{tab:scaling_form} varies budgets with the same backbone, decoder, action horizon, splits, and observations.

\paragraph{Trainable parameters and resource cost.}
Simulation training uses eight NVIDIA A100 80GB GPUs and BF16 mixed precision.
Base backbone weights are frozen; rank-32 LoRA adapters target all linear modules, while the action head and MPCoT modules are trainable.
Training uses $(K,M)=(3,2)$. Within each benchmark setting, all latent configurations in Table~\ref{tab:scaling_form} share one checkpoint.
Under this setup, OpenVLA-OFT and MPCoT take 24 and 36 hours, respectively, with peak memory of 62 and 71 GB per GPU.
The MPCoT total includes candidate decoding, latent refinement, and reward evaluation.

\paragraph{Evaluation records.}
LIBERO results average 10 repeated evaluations of the same checkpoint trained with one seed. Each suite comprises 10 tasks and 50 episodes per task per evaluation ($10\times50\times10=5{,}000$ episodes).
Logs record task, episode, seed, success, and cumulative success rate.
They support episode-level uncertainty estimates for a fixed policy, but not variability across independently trained checkpoints.

\begin{table*}[t]
\centering
\normdbltable
\setlength{\tabcolsep}{3pt}
\caption{Comparison on CALVIN ABC$\rightarrow$D benchmark. Columns 1-step to 5-step report success rate (SR, \%), and Avg. len reports average successful sequence length.}
\label{tab:calvin_main}
\begin{tabular*}{\textwidth}{@{\extracolsep{\fill}}lcccccc@{}}
\toprule
Method & 1-step SR (\%) & 2-step SR (\%) & 3-step SR (\%) & 4-step SR (\%) & 5-step SR (\%) & Avg. len \\
\midrule
OpenVLA~\cite{openvla} & 91.3 & 77.8 & 62.0 & 52.1 & 43.5 & 3.27 \\
UniVLA~\cite{univla} & 95.5 & 85.8 & 75.4 & 66.9 & 56.5 & 3.80 \\
UnifiedVLA~\cite{unifiedvla} & 98.9 & 94.8 & 89.0 & 82.8 & 75.1 & 4.41 \\
OpenVLA-OFT~\cite{openvla_oft} & 96.9 & 92.0 & 85.7 & 80.4 & 72.9 & 4.28 \\
FLOWER~\cite{flower_vla} & 99.4 & 95.8 & 90.7 & 84.9 & 77.8 & 4.49 \\
VLA-Adapter~\cite{vla_adapter} & 99.1 & 94.6 & 88.8 & 82.8 & 76.5 & 4.42 \\
Seer~\cite{seer} & 96.3 & 91.6 & 86.1 & 80.3 & 74.0 & 4.28 \\
AVA-VLA~\cite{ava_vla} & 99.6 & 97.6 & 94.1 & 89.9 & 84.1 & 4.65 \\
\textbf{MPCoT (Ours)} & \textbf{99.8} & \textbf{98.9} & \textbf{96.8} & \textbf{93.7} & \textbf{89.4} & \textbf{4.79} \\
\bottomrule
\end{tabular*}
\end{table*}

\begin{table*}[t]
\centering
\normdbltable
\caption{ALOHA Mini full-task success over 30 trials per task and method. Counts are successes/trials; SR is in percent, and $\Delta$ denotes percentage-point gain.}
\label{tab:real_world}
\begin{tabular*}{\textwidth}{@{\extracolsep{\fill}}lccccc@{}}
\toprule
Task & OpenVLA-OFT trials & SR (\%) & MPCoT trials & SR (\%) & $\Delta$ \\
\midrule
Single-Object Pick-and-Place & 27/30 & 90.0 & 29/30 & \textbf{96.7} & \textbf{+6.7} \\
Language-Conditioned Pick-and-Place & 24/30 & 80.0 & 25/30 & \textbf{83.3} & \textbf{+3.3} \\
Multi-Object Sequential Manipulation & 20/30 & 66.7 & 24/30 & \textbf{80.0} & \textbf{+13.3} \\
Towel Folding & 17/30 & 56.7 & 21/30 & \textbf{70.0} & \textbf{+13.3} \\
Block Stacking & 19/30 & 63.3 & 24/30 & \textbf{80.0} & \textbf{+16.7} \\
\midrule
\textbf{Average / Total} & \textbf{107/150} & \textbf{71.3} & \textbf{123/150} & \textbf{82.0} & \textbf{+10.7} \\
\bottomrule
\end{tabular*}
\end{table*}

\paragraph{Metrics.}
For LIBERO, we report suite-level and average success rates (SR).
For CALVIN ABC$\rightarrow$D, we report 1--5-step SR and average successful sequence length.
Path Consistency measures agreement between the scorer-preferred path and the highest-return path:
\begin{equation}
\begin{aligned}
\mathrm{PathConsistency}
&=\frac{1}{N}\sum_{i=1}^{N}\mathbf{1}\!\left[\hat m_i^{s}=\hat m_i^{R}\right],\\
\hat m_i^{s}&=\operatorname*{arg\,max}_{m}s_{i,m},\quad
\hat m_i^{R}=\operatorname*{arg\,max}_{m}R_{i,m}.
\end{aligned}
\label{eq:path_consistency}
\end{equation}
Mechanism ablations additionally report path similarity and run-to-run standard deviation.

\paragraph{Real-world setup.}
ALOHA Mini uses three 224$\times$224 RGB views, language, and proprioception.
Both methods predict 25-step, 14-D absolute joint-action chunks, executing all steps at 25 Hz before replanning.
MPCoT trains at $(3,2)$ and evaluates at $(5,4)$; Sec.~\ref{sec:path_preference} defines its real-robot reward interfaces.

\subsection{Main Benchmark Results on LIBERO}
\label{sec:libero_results}
\textbf{MPCoT (Ours)} denotes \textbf{OpenVLA-OFT + MPCoT} throughout.
Table~\ref{tab:libero} separates joint all-suite and suite-specific training; comparisons within each protocol share observations, language, proprioception, and the 8-step action interface.
MPCoT adds refinement and training-only path supervision, with no extra evaluation observations or reward labels.
Average and Long SR improve in both settings, although near-ceiling performance limits these gains; CALVIN provides a harder sequential test.

\subsection{Long-Horizon Generalization on CALVIN ABC$\rightarrow$D}
\label{sec:calvin_results}
Table~\ref{tab:calvin_main} evaluates policies trained on A/B/C and tested on held-out D.
MPCoT achieves the highest reported SR at every sequence length, with a widening margin over OpenVLA-OFT for longer chains.
This supports improved sequential execution under cross-environment generalization; the ablations below examine individual component contributions.

\subsection{Real-World Evaluation}
\label{sec:real_world}
Table~\ref{tab:real_world} reports results on five ALOHA Mini tasks.
Both methods use the same training data and evaluation protocol, with 30 trials per task (150 per method).
We report full-task SR, excluding partial completion.
The multi-object task requires all five objects in the target box; towel folding requires completion of the prescribed folding sequence.

MPCoT improves average real-world SR from 71.3\% to 82.0\%, corresponding to 16 additional completed trials across the five tasks.
Gains are larger on multi-object manipulation, towel folding, and stacking than on the two shorter pick-and-place tasks.
This pattern is consistent with the CALVIN results, although task differences prevent attributing the gap solely to horizon length.
Supplementary videos illustrate repeated unsuccessful placement attempts and folding misalignment in OpenVLA-OFT, alongside completed MPCoT executions.

\paragraph{Policy-query latency.}
OpenVLA-OFT and MPCoT average 364 and 465 ms per query (+101 ms) over 100 queries on the same NVIDIA A100 GPU.
These times use native 25-step, 14-D joint actions; Table~\ref{tab:scaling_form} uses 8-step simulation actions.

\begin{table*}[t]
\centering
\normdbltable
\setlength{\tabcolsep}{2pt}
\renewcommand{\arraystretch}{1.05}
\caption{Depth--width scaling. SRs: \% (CALVIN: cumulative); latency: Sec.~\ref{sec:test_time_scaling}. $\dagger$: default; bold: best SR.}
\label{tab:ablate_km}
\label{tab:scaling_form}
\begin{tabular*}{\textwidth}{@{\extracolsep{\fill}}lccccccc@{}}
\toprule
Variant & $(K,M)$ & Agg. & \shortstack{LIBERO\\Avg. SR} & \shortstack{LIBERO\\Long SR} & \shortstack{CALVIN\\3-step SR} & \shortstack{CALVIN\\5-step SR} & \shortstack{Latency\\(ms)} \\
\midrule
OpenVLA-OFT (Direct) & (0,1) & -- & 96.8 & 95.3 & 85.7 & 72.9 & 24 \\
Latent CoT & (1,1) & -- & 98.2 & 97.9 & 94.8 & 82.5 & 26 \\
Latent CoT & (3,1) & -- & 98.4 & 98.1 & 95.6 & 88.3 & 29 \\
Latent CoT & (5,1) & -- & 98.6 & 98.4 & 96.3 & 88.9 & 33 \\
Latent CoT & (7,1) & -- & 98.7 & 98.5 & 96.5 & 89.1 & 38 \\
Multi-path Latent CoT & (1,3) & Soft & 98.3 & 98.1 & 95.4 & 84.3 & 27 \\
Multi-path Latent CoT & (3,2) & Soft & 98.5 & 98.3 & 95.9 & 88.6 & 31 \\
Multi-path Latent CoT & (3,4) & Soft & 98.7 & 98.6 & 96.5 & 89.1 & 34 \\
\textbf{Multi-path Latent CoT}$^\dagger$ & (5,4) & Soft & 98.9 & 98.9 & 96.8 & 89.4 & 38 \\
Multi-path Latent CoT & (5,6) & Soft & 98.9 & \textbf{99.0} & 96.8 & 89.4 & 43 \\
Multi-path Latent CoT & (7,6) & Soft & \textbf{99.0} & \textbf{99.0} & \textbf{96.9} & \textbf{89.5} & 49 \\
Textual CoT + Policy & -- & -- & 98.5 & 98.2 & 95.7 & 88.4 & 135 \\
\bottomrule
\end{tabular*}
\end{table*}

\begin{table*}[t]
\centering
\normdbltable
\caption{Multi-path reasoning mechanism ablation. Lower similarity indicates more diverse hypotheses; Std reports run-to-run variation.}
\label{tab:ablate_mechanism}
\begin{tabular*}{\textwidth}{@{\extracolsep{\fill}}p{0.35\textwidth}cccccc@{}}
\toprule
Variant & Soft & Div. & Sim. $\downarrow$ & Avg. (\%) & Long (\%) & Std $\downarrow$ \\
\midrule
Single-path Latent CoT ($K{=}3,M{=}1$) & No & No & -- & 98.4 & 98.1 & 0.52 \\
Multi-path (Hard argmax) & No & No & 0.90 & 98.5 & 98.2 & 0.84 \\
Multi-path (Uniform avg) & Yes & No & 0.83 & 98.6 & 98.4 & 0.53 \\
Multi-path (Soft, no reg) & Yes & No & 0.91 & \textbf{98.7} & \textbf{98.6} & 0.41 \\
Multi-path (Ours, $K{=}3,M{=}4$) & Yes & Yes & \textbf{0.66} & \textbf{98.7} & \textbf{98.6} & \textbf{0.18} \\
\bottomrule
\end{tabular*}
\end{table*}

\begin{table*}[t]
\centering
\normdbltable
\setlength{\tabcolsep}{3pt}
\caption{Reward-guided path preference ablation under the same protocol as the main results. Avg and Long denote LIBERO success rates. Path Consistency measures agreement between the scorer-preferred path and the highest-return path.}
\label{tab:reward_main_ablation}
\begin{tabular*}{\textwidth}{@{\extracolsep{\fill}}llccccc@{}}
\toprule
Variant & Rewards & Avg & Long & CALVIN 3-step & CALVIN 4-step & Path Cons. (\%) \\
\midrule
w/o Reward & BC only & 98.4 & 97.6 & 94.7 & 90.8 & 68.5 \\
Action Only & $r^{\mathrm{act}}$ & 98.6 & 98.0 & 95.3 & 91.7 & 73.2 \\
Progress Only & $r^{\mathrm{prog}}$ & 98.6 & 98.2 & 95.6 & 92.1 & 74.8 \\
Endpoint Only & $r^{\mathrm{end}}$ & 98.5 & 98.1 & 95.1 & 91.5 & 72.4 \\
Action + Progress & $r^{\mathrm{act}}+r^{\mathrm{prog}}$ & 98.7 & 98.5 & 96.1 & 92.8 & 79.6 \\
Action + Endpoint & $r^{\mathrm{act}}+r^{\mathrm{end}}$ & 98.7 & 98.4 & 95.9 & 92.6 & 78.9 \\
Progress + Endpoint & $r^{\mathrm{prog}}+r^{\mathrm{end}}$ & 98.8 & 98.6 & 96.3 & 93.1 & 81.2 \\
\textbf{Full Reward (Ours)} & $\mathbf{r^{\mathrm{act}}+r^{\mathrm{prog}}+r^{\mathrm{end}}}$ & \textbf{98.9} & \textbf{98.9} & \textbf{96.8} & \textbf{93.7} & \textbf{84.3} \\
\bottomrule
\end{tabular*}
\end{table*}

\subsection{Latent Reasoning Structure Ablations}
\label{sec:latent_structure_ablation}
With $M=1$, increasing $K$ from 1 to 5 raises CALVIN 3-step SR from 94.8\% to 96.3\% and 5-step SR from 82.5\% to 88.9\%.
Moving to $K=7$ adds 0.2 points on each metric for another 5 ms.

Width also improves shallow refinement: at $K=1$, increasing $M$ from 1 to 3 raises CALVIN 3-/5-step SR from 94.8/82.5\% to 95.4/84.3\%, with latency increasing from 26 to 27 ms.
At $K=3$, the $M=1,2,4$ sweep raises CALVIN 5-step SR from 88.3\% to 88.6\% and 89.1\%, respectively.
At $K=5$, increasing $M$ from 1 to 4 improves LIBERO-Long and CALVIN 5-step SR by 0.5 percentage points at 5 ms additional latency (Table~\ref{tab:ablate_km}).
At the same reported 38 ms, $(5,4)$ exceeds $(7,1)$ by 0.4 points on LIBERO-Long and 0.3 on both CALVIN metrics.
Width thus contributes beyond extra single-path depth at this latency.

In Table~\ref{tab:ablate_mechanism}, learned soft weights outperform hard selection and uniform averaging.
Adding diversity regularization reduces path similarity from 0.91 to 0.66 and reported standard deviation from 0.41 to 0.18, while average and Long SR remain unchanged.

\subsection{Test-Time Scaling and Reasoning Efficiency}
\label{sec:test_time_scaling}
Table~\ref{tab:scaling_form} compares success and latency using one checkpoint per benchmark setting, varying inference budgets without retraining.

\paragraph{Textual-CoT configuration.}
The baseline uses the same OpenVLA-OFT backbone and LIBERO data, without additional CoT, rationale, or robot data. Prompt:
\begin{quote}
\small
Given the current observation and task instruction, briefly describe the current state, identify the remaining subgoal, and reason about the next action. Keep the reasoning concise. Then predict the robot action.
\end{quote}
Greedy decoding (temperature 0) generates at most 32 reasoning tokens (approximately 20 on average). The same backbone re-encodes the instruction with this reasoning before the original action head predicts actions.

\paragraph{Latency protocol.}
Timing uses an NVIDIA A100 80GB, BF16, batch size 1, 20 warm-ups, and 100 iterations with CUDA synchronization before and after measurement.
Textual-CoT timing includes image encoding, text generation, re-encoding, and action decoding. These policy-query times exclude robot execution.

Larger budgets yield diminishing returns: $(5,6)$ preserves both CALVIN SRs at 43 versus 38 ms; $(7,6)$ improves each reported SR by 0.1 percentage points over $(5,4)$ at 49 ms (+28.9\% latency).
We retain $(5,4)$ as the default accuracy--latency trade-off: 89.4\% 5-step SR at 38 ms versus Textual CoT's 88.4\% at 135 ms. This comparison includes re-encoding overhead; parameters and compute are not jointly matched.

\subsection{Reward-Guided Path Preference Supervision}
\label{sec:ablate_opt_path}
Table~\ref{tab:reward_main_ablation} holds architecture and inference compute fixed.
Each reward improves over behavior cloning alone; the full objective outperforms individual signals and pairwise combinations.
Adding progress to action-plus-endpoint supervision raises CALVIN 4-step SR from 92.6\% to 93.7\%, supporting its complementary contribution.
Consistency rewards measure expert alignment, while progress predicts completion.
Path Consistency measures scorer--return agreement, not independent physical task success.

\section{CONCLUSION}
MPCoT allocates VLA inference compute through latent refinement depth $K$ and hypothesis width $M$, preserving the action interface without reasoning tokens or inference-time rewards.
Training-only reward supervision guides hypothesis aggregation.
LIBERO, CALVIN, and five ALOHA Mini tasks show improved execution; ablations support refinement, aggregation, diversity, and path-preference learning.
The measured accuracy--latency trade-off favors $(5,4)$ as a practical default, with marginal gains at larger budgets.
This configuration retains the original action interface and improves real-world full-task success by 10.7 percentage points over OpenVLA-OFT.

\section{LIMITATIONS}
Evaluation uses a single training seed, leaving checkpoint variability unmeasured.
Real-world tests use one embodiment and 30 trials per task per method in controlled tabletop conditions; robustness across embodiments, cameras, and unconstrained environments remains untested.
Qwen3-VL label bias may propagate to the scorer; independent progress-score validation and evaluator-sensitivity studies remain outside this evaluation.
Component ablations also do not replace a baseline jointly matched for parameters and compute.
Future work will examine hardware transfer, evaluator sensitivity, and stricter matched controls.
We will release reward code, prompts, configurations, and evaluation scripts to support reproduction.

\bibliographystyle{IEEEtran}
\bibliography{refs}

\end{document}